# Robust Metric Learning based on the Rescaled Hinge Loss

Sumia Abdulhussien Razooqi Al-Obaidi[1], Davood Zabihzadeh*[2], Hamideh Hajiabadi[3]

[1] Supervision and Scientific. Evaluation Apparatus, Ministry of Higher Education and Scientific Research, Baghdad, Iraq

[2] Computer Department, Engineering Faculty, Sabzevar University of New Technology, Sabzevar, IRAN

[3] Department of Computer Engineering, Birjand University of Technology, Birjand, Iran

* Corresponding Author

sumia.abedai@mohesr.gov.iq, d.zabihzadeh@mail.um.ac.ir, hajiabadi@birjandut.ac.ir

## Abstract

Distance/Similarity learning is a fundamental problem in machine learning. For example, kNN classifier or clustering methods are based on a distance/similarity measure. Metric learning algorithms enhance the efficiency of these methods by learning an optimal distance function from data. Most metric learning methods need training information in the form of pair or triplet sets. Nowadays, this training information often is obtained from the Internet via crowdsourcing methods. Therefore, this information may contain label noise or outliers leading to the poor performance of the learned metric. It is even possible that the learned metric functions perform worse than the general metrics such as Euclidean distance. To address this challenge, this paper presents a new robust metric learning method based on the Rescaled Hinge loss. This loss function is a general case of the popular Hinge loss and initially introduced in (Xu et al. 2017) to develop a new robust SVM algorithm. In this paper, we formulate the metric learning problem using the Rescaled Hinge loss function and then develop an efficient algorithm based on HQ (Half-Quadratic) to solve the problem. Experimental results on a variety of both real and synthetic datasets confirm that our new robust algorithm considerably outperforms state-of-the-art metric learning methods in the presence of label noise and outliers.

## 1. Introduction

Similarity/Distance measures are a key component in many machine learning and data mining algorithms. For example, clustering methods or kNN classifier are based on a similarity/distance measure. In addition, information retrieval systems require a measure to sort the retrieved objects based on degrees of relevancy to a query object. However, standard measures such as Euclidean distance or cosine similarity are not appropriate for many applications. For example in Figure 1, the $w_1$ feature, unlike $w_2$, is useful to discriminate data for the classification task while standard measures assign the same weight to both of these features.

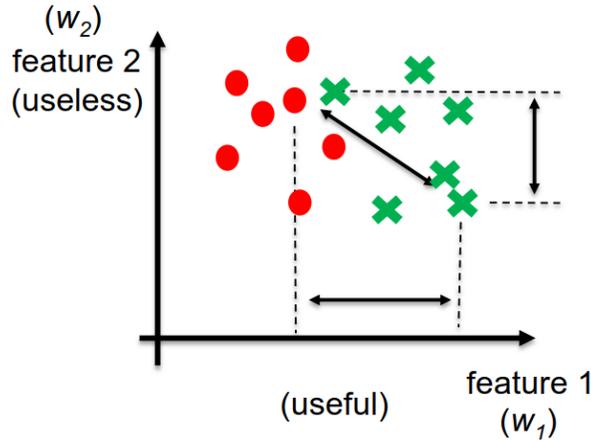

Figure 1: The problem of Standard Similarity/Distance measures. These measures assign the same importance to both $w_1$ and $w_2$ features while $w_1$ is more useful than $w_2$ to discriminate date items for the classification task. The image downloaded from Google.

Metric learning learns a distance function from data which brings conceptually related data items together while keeps unrelated ones at a distance. Distance Metric Learning (DML) algorithms often require training information in the form of pair or triplet side information as follows:

$S = \{(x_i, x_j) \mid x_i \text{ and } x_j \text{ are similar}\}$

$D = \{(x_i, x_j) \mid x_i \text{ and } x_j \text{ are dissimilar}\}$

$T = \{(x_i, x_i^+, x_i^-) \mid x_i \text{ should be more closer to } x_i^+ \text{ than to } x_i^-\}$

DML successfully applied in many applications such as Content Based Information Retrieval (CBIR) (Chechik et al. 2010; Hao et al. 2014; Li et al. 2017; Wu et al. 2016), Person re-identification (Bak and Carr 2017; Lin et al. 2017), Visual Tracking (Jiang et al. 2012), and Image Annotation (Guillaumin et al. 2009).

Despite achievements of DML in various applications, DML algorithms often are sensitive to outliers which may lead to substantial performance degradation. To address this challenge, Robust DML techniques are developed. These techniques can be classified into three categories

1 - Regularization-based,

2 - Probabilistic,

3 - Methods with robust loss function

A regularization-based approach often helps to avoid over-fitting in a small training dataset or when it contains noisy features. However, this approach is less effective in the presence of label noise.

Probabilistic methods estimate the probability of label noise for the input data. These methods often have non-convex formulation and are computationally expensive.

Another type of methods utilizes robust loss functions against outliers which has not been receiving much attention so far. Many DML algorithms (Chechik et al. 2010; Hao et al. 2014; Shi et al. 2014; Wang et al. 2015; Weinberger and Saul 2009; Wu et al. 2016; Yang et al. 2010) use the margin-based Hinge loss. These algorithms mainly use triplet side information to learn the metric. Since triplet side information has more emphasis on relative distance instead of the absolute one, these algorithms usually have higher performance compared to those which utilize the pairwise constraints. However, the unboundedness of the Hinge function causes outliers to have a large loss which leads to the poor performance of the DML algorithm in noisy environments.

To address this issue, this work proposes a robust metric learning method based on the Rescaled Hinge loss. We initially formulate the metric learning problem using the Rescaled Hinge loss and then provide an efficient algorithm based on HQ (Half-Quadratic) {Geman, 1995 #127} to solve the problem. The proposed approach is rather general and easily can be applied to any DML algorithm based on the Hinge loss.

The rest of the paper is organized as follows: Section 2 reviews related works. Section 3 presents the formulation of metric learning problem using the Rescaled Hinge loss as well as the development of the proposed algorithm. A comparison with state-of-the-art methods and experimental results are presented in Section 4. Finally, Section 5 concludes with remarks and recommendations for future work.

## 2. Related Work

Nowadays it is common to use crowdsourcing or similar techniques to harvest data from the Internet, particularly in the field of computer vision and machine learning (Krishna et al. 2016). Hence, one emerging challenge in DML is noisy side information. Robust DML algorithms are developed to address this problem.

Many metric learning algorithms add a regularizer to the objective function. The optimization problem of DML typically has the following general form:

$$\min_{M} l(M, C) + \lambda r(M) \tag{1}$$

where $M$ is the metric, $C$ is a set of input side information and $r(M)$ is a regularizer on the parameters of metric. Seminal regularization term include *Frobenius norm* (Chechik et al. 2010; Hao et al. 2014; Huang et al. 2012; Jin et al. 2009; Nguyen et al. 2017; Wu et al. 2016) *trace* (Niu et al. 2014; Shen et al. 2012) and *logdet* (Davis et al. 2007; Jain et al. 2012). Although a regularizer helps to avoid over-fitting in a small or contaminated with noisy features dataset, this technique is less effective in the presence of label noise.

Most research in Robust DML is dedicated to probabilistic methods. These methods usually have non-convex formulation and are computationally expensive. In the following, we discuss some prominent methods in this domain.

GMENs[1](Yang et al. 2010) proposes a framework for learning from noisy side information based on the generalized maximum entropy model. GMENs initially formulates the DML as a binary classification task. Then, it extends the problem to the case of noisy side information. The authors provide theoretical analysis which indicates that under a certain assumption, the solution found by GMENS in noisy environment converges to the one obtained from perfect side information.

---

[1] Generalized Maximum Entropy Model for learning from Noisy side information

RML (Huang et al. 2012) initially formulates the DML task as a combinatorial integer optimization problem. It assumes a priori knowledge that at most $1 - \eta$ percent of triplet constraints is noisy, exists. learning task is then formulates as:

$$\min_{t, M \geqslant 0} t + \frac{\lambda}{2} \|M\|_F^2$$
$$s.t. \quad t \geq \sum_{i=1}^{N} q_i l\left(d_M^2(x_i, x_i^+) - d_M^2(x_i, x_i^-)\right), \quad \forall q \in Q(\eta) \tag{2}$$

where $\{(x_i, x_i^+, x_i^-) \mid i = 1, 2, \ldots, N\}$ is the set of triplet constraints, $\|M\|_F$ is the Frobenius norm regularizer and the set $Q(\eta)$ is defined as: $Q(\eta) = \{q \in [0,1]^N : \sum_{i=1}^{N} q_i \leq N\eta\}$.

The above problem is transformed to a semi-infinite programming problem [4] and further to a convex optimization problem. the final problem is solved by Nesterov's smooth optimization method [7] which converges considerably faster than the sub-gradient method.

RNCA extends a previous non-parametric DML algorithm, i.e., NCA. It analyzes the effect of label noise on the derivative of likelihood with respect to the metric. Afterwards, RNCA proposes to model the probability of the true label of each point so as to reduce that effect. The model is then optimized within the EM framework.

Recently, BLMNN[1] (Wang and Tan 2018) proposes metric learning using Bayesian inference. Instead of the point estimation of the distance matrix, BLMNN estimates the posterior distribution of the metric using SVI (Stochastic Variational Inference) which decreases over-fitting problem in the small or noisy training set. This method extends the popular LMNN algorithm for the Bayesian learning scheme. Let S indicate the training triplet set, $p(M)$ is the *prior* distribution on the metric $M$, and $p(S|M)$ denote the *likelihood* defined as follows:

$$p(S|M) = \prod_{(x_i, x_i^+, x_i^-) \in S} p(x_i, x_i^+, x_i^- | M)$$
$$= C \prod_{(x_i, x_i^+, x_i^-) \in S} \exp\{-2 \max(1 + d_M^2(x_i, x_i^+) - d_M^2(x_i, x_i^-), 0)\} \tag{3}$$

then BLMN provide an algorithm to estimate the *posterior* $p(M|S)$. Both the theoretical analysis and experimental results show that any outlier has a bounded influence on the learned model.

Some methods utilizes robust loss functions against outliers. For instance, (Wang et al. 2014) use the $l_1$ norm to deal with noisy features and examples. Since the introduced objective function is very sensitive to the initial value of the solution, this method uses (Xiang et al. 2008) to initialize the metric. Also, the unbounded $l_1$ norm function has linear growth versus the error caused by an outlier. Hence, this method is not robust against label noise.

The proposed algorithm is based on HQ minimization method that is a fast alternating direction method. Suppose $F(M)$ is a non-quadratic function of $M$. The main idea of HQ is to introduce an auxiliary variable $v$ using conjugate theory {Boyd, 2004 #48} and construct a new cost function $F(M, v)$ which is quadratic in $M$ and such that:

$$(M^*, v^*) = \underset{M,v}{\arg\min}\, F(M, v) \rightarrow M^* = \underset{M}{\arg\min}\, F(M)$$

---

[1] Bayesian Large Margin Nearest Neighbor

## 3. Proposed Method

As mentioned, many DML algorithms use the Hinge loss function ($l_{hinge}$) shown in Figure 2. These methods typically solve an optimization problem with the following general form:

$$\underset{M}{\text{minimize}}\, \text{reg}(M) + C \sum_{i=1}^{|\mathcal{T}|} l_{hinge}(x_i, x_i^+, x_i^-) \tag{4}$$

where $M$ indicate the metric, $\text{reg}(M)$ is a regularizer function and $\mathcal{T}$ is the input triplet side information. The Hinge loss on the training triplet $(x_i, x_i^+, x_i^-)$ is defined as:

$$l_{hinge}(x_i, x_i^+, x_i^-) = \max\left(0, 1 + d_M^2(x_i, x_i^+) - d_M^2(x_i, x_i^-)\right) \tag{5}$$

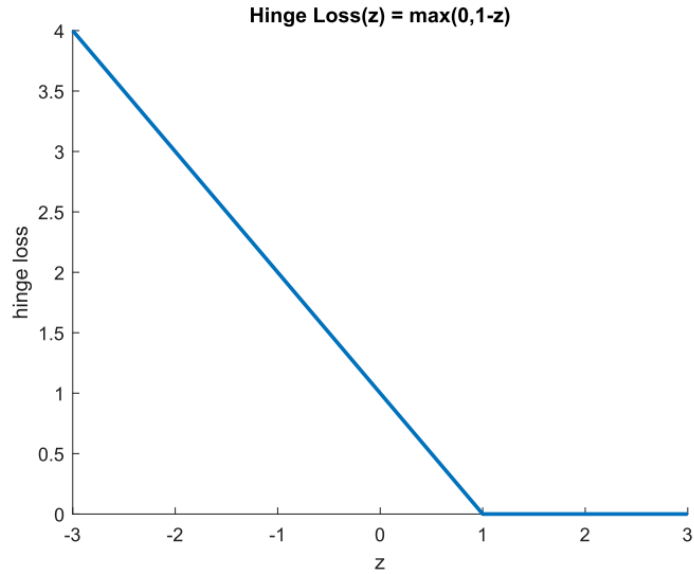

**Figure 2: The margin-based Hinge loss function.**

Let $z_i = d_M^2(x_i, x_i^-) - d_M^2(x_i, x_i^+)$, then the optimization problem (4) can be rewritten as:

$$\underset{M \succcurlyeq 0}{\text{minimize}}\, \text{reg}(M) + C \sum_{i=1}^{|\mathcal{T}|} l_{hinge}(z_i) \tag{6}$$

where $l_{hinge}(z) = \max(0, 1 - z)$

As shown in Figure 2, $l_{hinge}$ limitlessly grows versus the variable $z$. The unboundedness of the $l_{hinge}$ function causes outliers to have a large loss which results in poor performance of the learned metric.

To address this challenge, this paper presents a novel robust DML algorithm with the Rescaled Hinge loss ($l_{rhinge}$). This loss function is initially introduced in (Xu et al. 2017) to develop a new robust SVM algorithm against outliers. $l_{rhinge}$ is defined as follows:

$$l_{rhinge}(z) = \beta \left[1 - \exp\left(-\eta l_{hinge}(z)\right)\right] \tag{7}$$

where $\eta$ is a rescaling factor and $\beta = 1/(1 - \exp(-\eta))$ is a normalizing constant which ensures that $l_{rhinge}(0) = 1$ (similar to the Hinge loss).

Figure 3 shows the diagram of the $l_{rhinge}$ loss with different values for $\eta$. As seen, the large losses incurred by outliers can be controlled by adjusting the parameter $\eta$. Also, (Xu et al. 2017) proves that the Hinge loss is a special case of $l_{rhinge}$. More precisely, $\lim_{\eta \to 0} l_{rhinge}(z) = l_{hinge}(z)$.

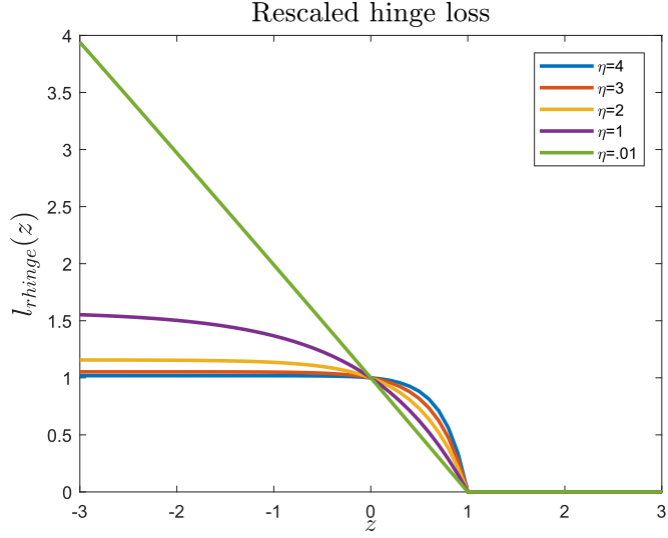

**Figure 3: The Rescaled hinge loss function with different $\eta$ values**

The objective function of the proposed robust DML method is formulated as follows:

$$\underset{M \geq 0}{\text{minimize}} \ reg(M) + C \sum_{i=1}^{|\mathcal{T}|} l_{rhinge}(z_i)$$

where  (8)

$$z_i = d_M^2(x_i, x_i^-) - d_M^2(x_i, x_i^+)$$

In the next subsection, we derive an efficient algorithm based on HQ to solve the above optimization problem.

### 3.1 RDML[1]

Let consider the regularizer term as $\frac{1}{2}\|M\|_F^2$, then we have

$$\underset{M \geq 0}{\text{minimize}} \ \frac{1}{2}\|M\|_F^2 + C \sum_{i=1}^{|\mathcal{T}|} l_{rhinge}(z_i) \tag{9}$$

By simple arithmetic modification, (9) can be rewritten as

$$\underset{M \succcurlyeq 0}{\text{maximize}} \quad -\frac{1}{2}\|M\|_F^2 + C\beta \sum_{i=1}^{|\mathcal{T}|} \exp\left(-\eta l_{hinge}(z_i)\right) \quad (10)$$

---

[1] Robust DML

As proved in (Xu et al. 2017), according to conjugate function theory

$$\exp\left(-\eta l_{hinge}(z)\right) = \sup_{v<0}\left(\eta l_{hinge}(z)v - g(v)\right) \quad (11)$$

where $g(v) = -v\log(-v) + v, \ (v < 0)$

According to relation (11):

$$\begin{aligned}
f_2(M) &= -\frac{1}{2}\|M\|_F^2 + C\beta \sum_{i=1}^{|\mathcal{T}|} \exp\left(-\eta l_{hinge}(z_i)\right) \\
&= -\frac{1}{2}\|M\|_F^2 + C\beta \sum_{i=1}^{|\mathcal{T}|} \sup_{v_i<0}\left(\eta l_{hinge}(z_i)v_i - g(v_i)\right) \\
&= -\frac{1}{2}\|M\|_F^2 + C\beta \sup_{v<0}\left(\sum_{i=1}^{|\mathcal{T}|}\left(\eta l_{hinge}(z_i)v_i - g(v_i)\right)\right) \\
&= \sup_{v<0}\left(-\frac{1}{2}\|M\|_F^2 + C\beta \sum_{i=1}^{|\mathcal{T}|}\left(\eta l_{hinge}(z_i)v_i - g(v_i)\right)\right)
\end{aligned} \quad (12)$$

where $\boldsymbol{v} = [v_1, v_2, \dots, v_N] < \boldsymbol{0} \ (N = |\mathcal{T}|)$. The second equation holds since the functions $\eta l_{hinge}(z_i)v_i - g(v_i) \ i = 1,2,\dots,N$ are independent in terms of $v_i$ and the last one establishes since $-\frac{1}{2}\|M\|_F^2$ is constant respect to $\boldsymbol{v}$. Substituting (12) into (10) yields:

$$\underset{M \succcurlyeq 0, v<0}{\text{maximize}} \ f_3(M,v) = -\frac{1}{2}\|M\|_F^2 + C\beta \sum_{i=1}^{|\mathcal{T}|}\left(\eta l_{hinge}(z_i)v_i - g(v_i)\right) \quad (13)$$

We solve the above problem by *alternating optimization* method. More precisely, given a current value of $\boldsymbol{v}$, (13) is optimized over $M$ and then given $M$, it is optimized over $\boldsymbol{v}$. We repeat this procedure until convergence.

Let the superscript $s$ denote the $s^{th}$ iteration, then if we optimize (13) just over $\boldsymbol{v}$, it will be equivalent to

$$\underset{v^s<0}{\text{maximize}} \ \sum_{i=1}^{|\mathcal{T}|}\left(\eta l_{hinge}(z_i)v_i - g(v_i)\right) \quad (14)$$

(14) has a closed-form solution as follows:

$$v_i^s = -\exp\left(-\eta l_{hinge}(z_i)\right), \quad i = 1,2,\dots,|\mathcal{T}| \quad (15)$$

After obtaining $\boldsymbol{v}^s$, we can find $M^{s+1}$ from the following optimization problem.

$$\underset{M \succcurlyeq 0}{\text{maximize}} \ -\frac{1}{2}\|M\|_F^2 + C\beta \sum_{i=1}^{|\mathcal{T}|}\left(\eta l_{hinge}(z_i)v_i\right) \quad (16)$$

The above optimization problem is equivalent to

$$\underset{M \succcurlyeq 0}{\text{minimize}} \, f(M) = \frac{1}{2} \|M\|_F^2 + \sum_{i=1}^{|\mathcal{T}|} C_i \xi_i \tag{17}$$

subject to

$$1 - d_M(x_i, x_i^-) + d_M(x_i, x_i^+) \leq \xi_i, \quad \xi_i \geq 0 \quad i = 1, 2, \ldots, |\mathcal{T}|$$

where $C_i = C\beta\eta(-v_i)$ indicate the weight of the $i^{th}$ triplet.

Since $f(M)$ is convex respect to $M$, we can use the stochastic sub-gradient method to find the global minimum of (17). The derivative of $f(M)$ with respect to $M$ is equal to

$$\frac{\delta f}{\delta M} = M + \sum_{i \in \mathcal{J}^s} C_i(1 - A^s) \tag{18}$$

where $A^s = (x_i - x_i^-)(x_i - x_i^-)^T + (x_i - x_i^+)(x_i - x_i^+)^T$

Here, $\mathcal{J}^s$ denote the set of active constraints in the current batch. According to (18), the gradient step to update $M$ is

$$\begin{aligned} M^{(new)} &= M^{(old)} - \lambda \left( M^{(old)} - \sum_{i \in \mathcal{J}^s} C_i(1 - A^s) \right) \\ &= (1 - \lambda)M^{(old)} - \lambda \sum_{i \in \mathcal{J}^s} C_i(1 - A^s) \end{aligned} \tag{19}$$

where $\lambda$ is the learning rate. Algorithm 1 summarizes the steps of RDML

---

**Algorithm1**. RDML (Robust Distance Metric Learning)

---

Input: $\lambda$: learning rate

    1. Initialize Matrix $M$ with Identity matrix

    2. Initialize weight vector $v$: $v = 1$

    3. for $s = 1, 2, \ldots, MaxHQIter$

        3.1. update weight vector $v$ from (15)

        3.2. **repeat**

            3.2.1. update matrix $M$ from (19)

        3.3. **until** convergence

        3.3. $M = psd(M)$

    end;

---

Here, the function $psd(M)$ project $M$ into the cone of p.s.d (positive.semi.definite) matrices.

Note that, the proposed approach is rather general and can be easily applied to any DML algorithm with the Hinge loss. Here, we simply use the Frobenius norm as a regularizer and the stochastic gradient method to optimize the metric.

### 3.2 Run Time Analysis

To analyze the time complexity of our model, let $T_h$ denote the time complexity of a DML algorithm based on the Hinge loss. If we apply our method to this algorithm, in addition of optimizing the metric, our method computes the vector $\boldsymbol{v}$ in each iteration according to equation (15). This computation requires $O(p|\mathcal{T}|)$ where $p \ll d$ is the rank of matrix $\boldsymbol{M}$ and $|\mathcal{T}|$ is the number of training triplets. Hence, the overall time complexity of our method is $O(MAXHQIter \times (T_h + p|\mathcal{T}|))$. In practice the vector $\boldsymbol{v}$ is computed very faster compared to the optimization of metric $\boldsymbol{M}$ with $O(d^2)$ parameters ($d$ is the dimension of input data). So, $|T| \ll T_h$. Moreover, in our experiments always $MAXHQIter \leq 3$. Therefore, our method can be efficiently applied to any DML algorithm based on the Hinge loss.

### 3.3 Convergence Analysis

Following the similar analysis in {Xu, 2017 #105}, the convergence property of the RDML algorithm can be established. According to (13):

$$f_3(\boldsymbol{M}, \boldsymbol{v}) \leq f_2(\boldsymbol{M}) = -\frac{1}{2}\|\boldsymbol{M}\|_F^2 + C\beta \sum_{i=1}^{|\mathcal{T}|} \exp\left(-\eta l_{hinge}(z_i)\right) \leq C\boldsymbol{\beta}|\mathcal{T}|$$

The second inequality holds since $\|\boldsymbol{M}\|_F^2 \geq 0$ and $\exp\left(-\eta l_{hinge}(z_i)\right) \leq 1$. Hence, $f_3(\boldsymbol{M}, \boldsymbol{v})$ is upper bounded. In addition, according to (14) and (16) we have

$$f_3(\boldsymbol{M}^s, \boldsymbol{v}^s) \leq f_3(\boldsymbol{M}^s, \boldsymbol{v}^{s+1}) \leq f_3(\boldsymbol{M}^{s+1}, \boldsymbol{v}^{s+1})$$

Consequently, the sequence $\{f_3(\boldsymbol{M}^s, \boldsymbol{v}^s), s = 1,2,...\}$ converges.

## 4. Experimental Results

In this section, we evaluate the performance of the proposed method on both synthetic and real datasets in presence of label noise. The results are compared with some peer DML methods.

In the first experiment, we investigate the ability of the proposed method to find outliers as well as reducing their effects on the learning process. For this purpose, a synthetic dataset with label noise is generated as shown in Figure 4(a). According to equation (17), $C_i = C\beta\eta(-v_i)$ indicates the weight of triplet $(x_i, x_i^+, x_i^-)$. We define the weight of an instance as the sum of weights of triplets in which it participates as the first element. In Figure 4(a), the normalized weight of each outlier is indicated. It is calculated by dividing the weight by the total weights. Note that, the normalized weights in this figure are multiplied by 1000 for a better representation. Figure 4(c) show the results after 3 iterations of RDML. As the results indicate, the weight of all instances with label noise is gradually decreased. The result confirm that RDML can effectively reduce the effects of outliers on the metric learning process.

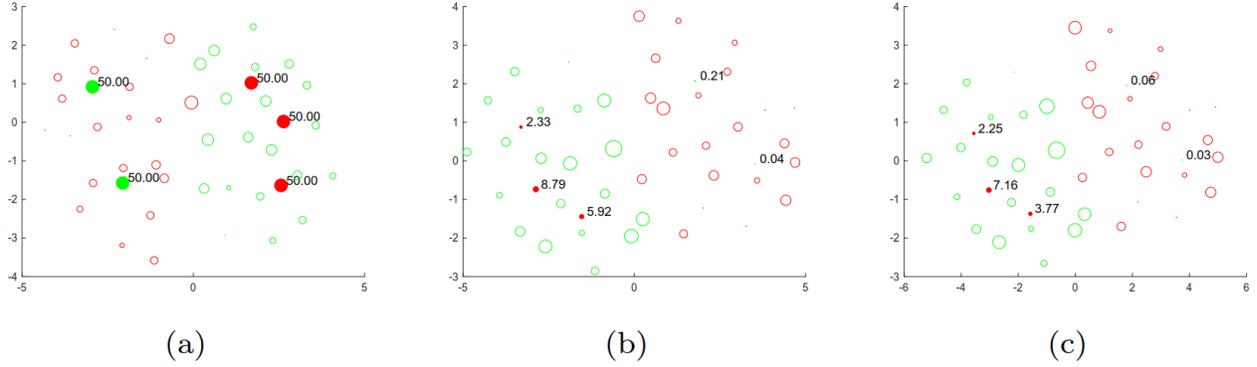

**Figure 4: (a) Initial normalized weights of outliers in the synthetic dataset. (b), (c) The normalized weights after 2 and 3 iterations of the RDML algorithm**

We also evaluate the performance of RDML versus outliers in a synthetic dataset with two classes sampled from two Gaussian distributions. The generation of the outliers is controlled by two parameters: $outlier\_ratio$ and $outlier\_intensity$. More precisely, we sample outliers in the $i^{th}$ class ($i = 1,2$) from $\mathcal{N}(x|\boldsymbol{\mu}_{oi}, \boldsymbol{\Sigma}_{oi})$ where

$$\boldsymbol{\mu}_{oi} = (1 - outlier\_intensity) \times \boldsymbol{\mu}_i + outlier\_intensity \times \boldsymbol{\mu}_j$$

$$\boldsymbol{\Sigma}_{oi} = 10 \times outlier\_intensity \times \boldsymbol{\Sigma}_i$$

Here, $\mathcal{N}(x|\boldsymbol{\mu}_i, \boldsymbol{\Sigma}_i)$ is a probability distribution of the $i^{th}$ class and $\boldsymbol{\mu}_j$ represents the mean of other class. Figure 5 shows the generated synthetic data where $outlier\_ratio = 0.2$ and $outlier\_intensity$ is changed from 0.1 to 0.3. The outliers can be identified by filled circles in this figure.

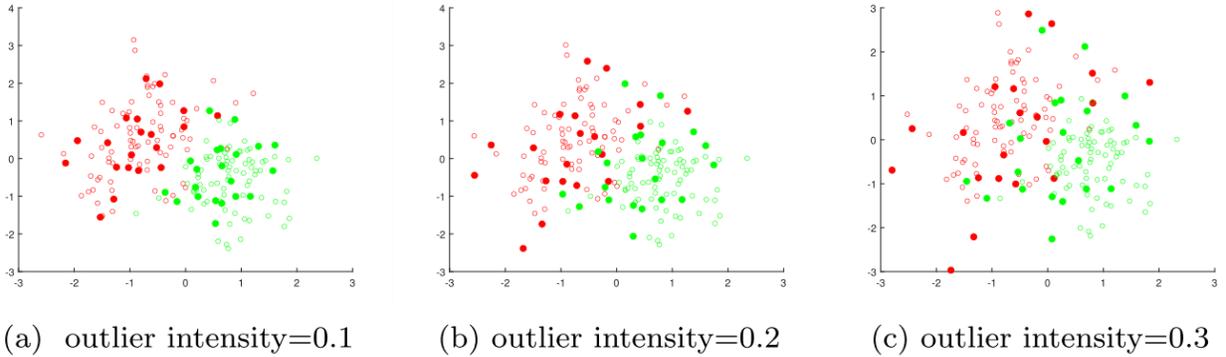

(a) outlier intensity=0.1    (b) outlier intensity=0.2    (c) outlier intensity=0.3

**Figure 5: Generated synthetic data where $outlier\_ratio = 0.2$ and $outlier\_intensity$ is changed from $0.1$ to $0.3$.**

In the first experiment, we study the effect of the increasing the ratio of outliers ($outlier\_ratio$) on the classification accuracy of kNN-RDML (kNN classifier with the metric learned by RDML). We changed the $outlier\_ratio$ from 0 to 40% and keep the $outlier\_intensity$ fixed equal to 0.3 . The results are plotted in Figure 12

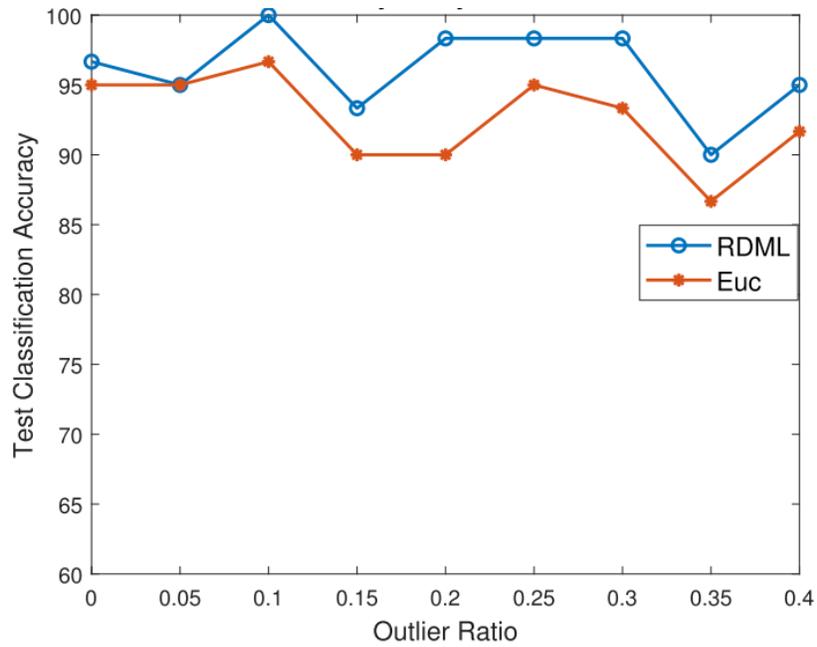

**Figure 6: Classification Accuracy of kNN using RDML and Euclidean metric on the synthetic data versus outlier ratio**

In the first experiment, we study the effect of the increasing ($outlier\_intensity$) on the classification accuracy of kNN-RDML. We changed the $outlier\_intensity$ from 0 to 0.70 and keep the $outlier\_ratio$ fixed equal to 0.3 . The results are depicted in Figure 7.

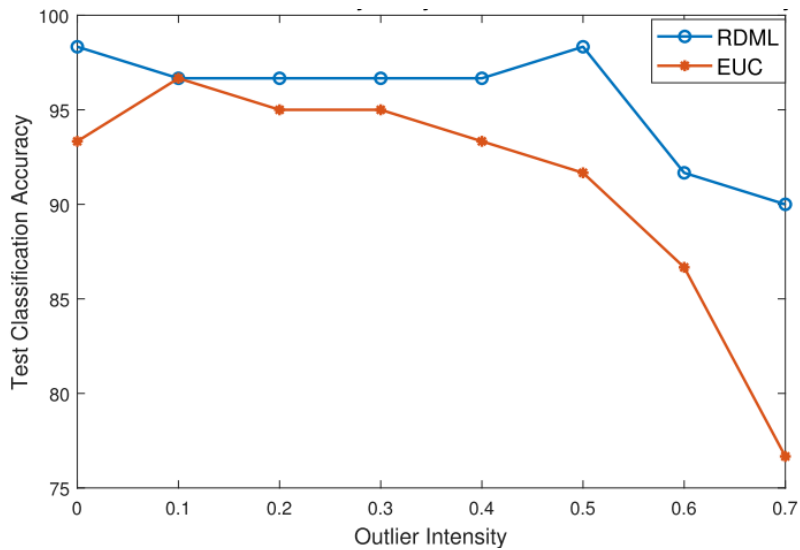

**Figure 7: Classification Accuracy of kNN using RDML and Euclidean metric on the synthetic data versus outlier intensity**

As the results in Figure 6 and Figure 7 indicate, the proposed method is robust against both the $outlier\_ratio$ and $outlier\_intensity$ and its performance is acceptable even when the large number of extreme outliers exists in the dataset.

In the next experiment, we evaluate the performance of the proposed method on real datasets contaminated with label noise. The results are compared with some peer DML methods. Table 1 reports the statistics of evaluated datasets in our experiments.

**Table 1-Statistics and explanations of evaluated datasets**

| Data Set | #classes | #samples | #dim | Feature extraction | d | Description |
|---|---|---|---|---|---|---|
| Wine (Lichman 2013) | 3 | 178 | 13 | None | 13 | Standard UCI classification dataset. https://archive.ics.uci.edu/ml/datasets/wine |
| Letters (Lichman 2013) | 26 | 20,000 | 16 | None | 16 | This dataset contains 20,000 examples of 26 English capital letters. Images of letters are generated from 20 different fonts. Then, 16 numerical attributes are extracted from these images. https://archive.ics.uci.edu/ml/datasets/letter+recognition |
| YaleFaceB (Lee et al. 2005) | 38 | 2,414 | 1,024 | PCA | 200 | This standard face recognition dataset includes 2,414 images of 38 classes. For each person, at most 64 images are taken under extreme illumination conditions. http://vision.ucsd.edu/~iskwak/ExtYaleDatabase/ExtYaleB.html |
| USPS(Hull 1994) | 10 | 9,289 | 256 | PCA | 100 | USPS is a handwritten digit dataset from envelopes by the U.S. Postal Service. The originally scanned digits are binary with different sizes and orientations. The images have been normalized which results in $16 \times 16$ grayscale images. USPS consists of 7291 training observations and 2007 test instances. https://www.kaggle.com/bistaumanga/usps-dataset |
| MNIST(LeCun et al. 1998) | 10 | 60,000 | 784 | PCA | 164 | MNIST is a popular handwritten digits dataset. It contains 10 categories (one per digit), 60,000 training examples and 10,000 test instances. The size of each image is $28 \times 28$ pixels. http://yann.lecun.com/exdb/mnist/ |
| Caltech101 (Li et al. 2004) | 101 | 9,146 | *variable* | CNN Resnet152 | 1000 | Caltech101 is a standard machine vision classification dataset. The images belong to 101 categories. Each class contains about 40 to 800 images and the size of each image is roughly $300 \times 200$ pixels. The images have no clutter and the objects often are centered in each image. http://www.vision.caltech.edu/Image_Datasets/Caltech101/ |

### 4.1 Experimental Setup

The data in the Wine, Letters, and YaleFaceB datasets are normalized so that the mean and variance of each attribute becomes 0 and 1 respectively. Also, to decrease the noise effect, PCA is utilized to reduce the dimension of images in the YaleFaceB, USPS and MNIST datasets. The parameter d in Table 1 denotes the input dimension after applying PCA.

In the Caltech101 dataset, the feature extraction is done using the Imagenet-resnet-152-dag deep network[1]. For this purpose, the images are initially rescaled to 224×224 and then subtracted from the mean image of the network. Subsequently, a total of 1000 features per image are extracted from the second last layer.

---

[1] downloaded from http://www.vlfeat.org/matconvnet/pretrained/

In the experiments, triplet side information is generated as follows. A training instance $x$ is set as similar to $k$'s nearest neighbors with the same label (*target neighbors*). Also, $x$ is considered as dissimilar to any *imposters*. These are any observations of a different class which violate the margin specified by $x$'s target neighbors. Finally, the triplet set is generated by the natural join of the similar and dissimilar sets. Figure 8 illustrates the concepts of *target neighbors* and *imposters* for a training example.

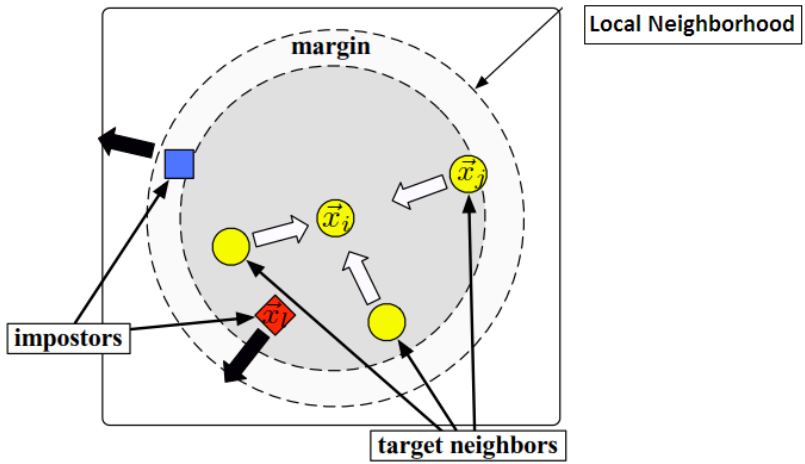

Figure 8: The illustration of the target neighbors and imposters of $x_i$. Reprinted from (Weinberger and Saul 2009)

We randomly split instances of datasets into training/test sets (70/30) at each run, except for USPS and MNIST which have predefined training/test sets. The results are compared with peer robust DML methods: BLMNN[1](Wang and Tan 2018), L1-DML[2] (Wang et al. 2014) and also with sparsity-based DML ones: SVM-trip[3] (Wang et al. 2015) and SCML-G[4] (Shi et al. 2014).

The hyperparameters of the competing methods are adjusted by 5-fold cross-validation as follows. The parameter $C$ in RDML and SVM-trip is selected from $\{.1, .5, 1, 3, 10\}$. The learning rate ($\lambda$) in RDML is chosen from the range $(10^{-6}, 10^{-2})$. Also, we select the number of bases in SCML-G from $\{400, 1000\}$, as recommended by its authors.

The kNN classifier $k = 3$ is adopted to evaluate the performance of the learned metrics. Figure 9 depicts the system flow of the proposed learning scheme which consists of two phases namely the *learning* and *test*.

---

[1] http://parnec.nuaa.edu.cn/xtan/data/BLMNN_demo.zip

[2] http://parnec.nuaa.edu.cn/xtan/data/BLMNN_demo.zip

[3] http://www.cs.cmu.edu/~deyum/Publications.htm

[4] http://mloss.org/software/view/553/

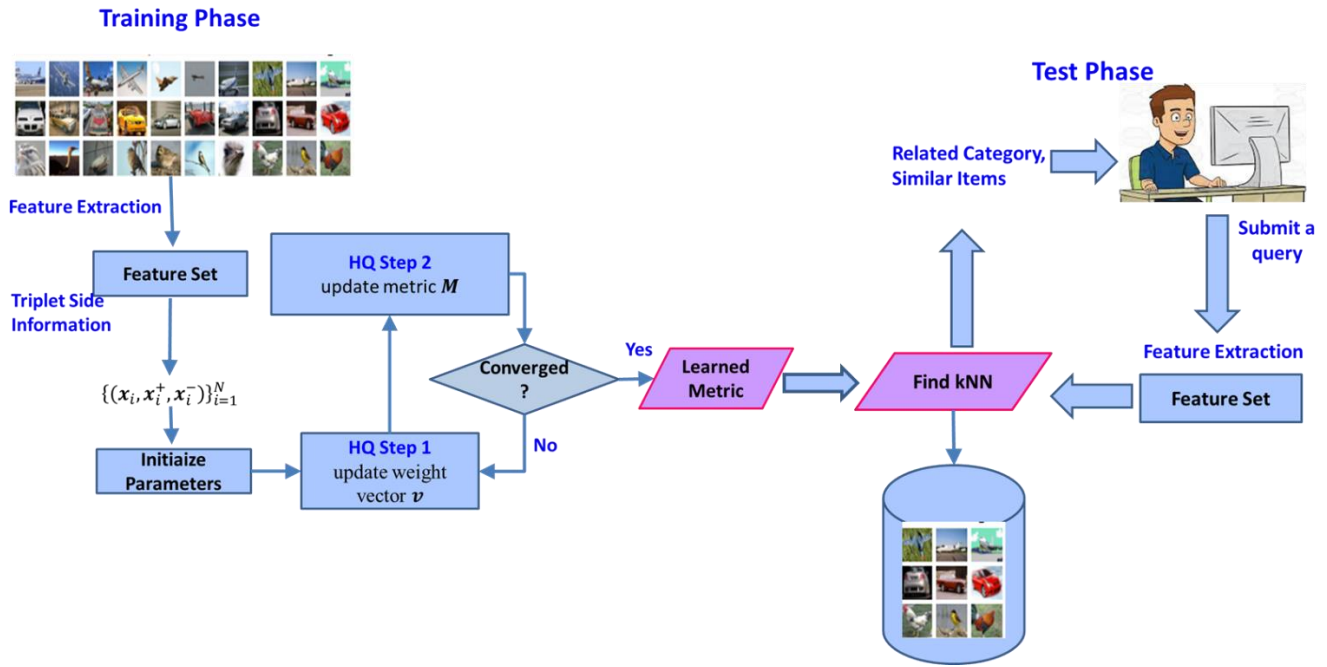

**Figure 9**: The system flow of the proposed metric learning scheme

## 4.2 Results and Analysis

The classification rate of the kNN classifier using the learned metric of the competing methods is reported in Table 2. Here, the parameter $nl$ indicates the percent of noise level in datasets. The results are obtained by averaging over 10 runs on these datasets. Also, Figure 7 depicts the accuracy of competing methods versus label noise (ranging from 0% to 20%). To make the comparison meaningful, we perform statistical analysis with p-value = 5% on the obtained results. The results, which have statistically significant differences are marked by * in Table 2. Also, these results are shown using box-plots in Figure 11. In addition, the running time of RDML and other DML methods are compared together on evaluated datasets and the results are shown in Figure 12.

**Table 2- The classification rate of the kNN classifier using the learned metric of the competing methods**

| Data Set | nl % | RDML | BLMNN | L1_DML | SCML_G | SVM_TRIP |
|---|---|---|---|---|---|---|
| Wine | 0 | 98.11±1.33 | 97.55±2.00 | 97.36±1.03 | **98.49±0.84** | 96.60±2.67 |
| | 5 | **97.74±2.46** | 96.60±2.19 | 96.98±1.03 | 96.60±2.07 | 94.34±3.77 |
| | 10 | **96.98±1.58** | 96.23±3.54 | 95.47±1.03 | 93.21±3.16 | 93.96±4.50 |
| | 20 | **95.09±3.91** | 92.83+-3.65 | 92.45±3.77 | 93.96±4.50 | 93.21±4.46 |
| Letters | 0 | **97.44±0.17*** | 94.10+-0.39 | 94.12±0.26 | 95.88±0.12 | 94.28±0.21 |
| | 5 | 93.68±0.20 | **93.79+-0.25** | 89.38±0.38 | 89.71±1.30 | 93.19±0.16 |
| | 10 | **92.02±0.45** | 91.51±0.55 | 87.84±0.42 | 88.58±0.99 | 91.69±0.14 |
| | 20 | **88.08±0.34** | 87.43±0.57 | 84.06±0.55 | 84.29±1.40 | 87.43±0.48 |
| YaleB32 | 0 | **96.72±0.84** | 90.51±0.97 | 90.21±0.50 | 95.86±1.00 | 87.54±1.04 |
| | 5 | **95.72±0.59*** | 86.73±1.59 | 87.37±0.80 | 93.30±0.41 | 86.48±0.62 |
| | 10 | **93.94±1.00*** | 87.90±1.90 | 86.40±0.87 | 92.71±0.40 | 84.87±1.13 |
| | 20 | **89.21±0.82*** | 83.56±1.32 | 81.53±2.08 | 86.43±0.45 | 80.81±1.50 |
| USPS | 0 | **95.45±0.03*** | 94.42±0.00 | 94.91±0.06 | 93.54±0.31 | 94.97±0.00 |
| | 5 | **94.90±0.22*** | 94.13±0.20 | 91.66±0.33 | 91.08±0.42 | 94.44±0.10 |
| | 10 | **93.80±0.29** | 92.75±0.47 | 90.11±0.28 | 90.60±0.92 | 93.56±0.20 |
| | 20 | **90.55±0.80** | 89.69±0.44 | 86.51±0.56 | 86.74±0.26 | 89.99±0.49 |
| MNIST | 0 | **97.84±0.08** | 97.77±0.00 | 97.09±0.01 | 97.45±0.06 | 97.40±0.00 |
| | 5 | **97.06±0.14** | 96.70±0.06 | 96.32±0.10 | 90.84±0.99 | 95.20±0.14 |
| | 10 | **95.59±0.34** | 95.53±0.13 | 95.08±0.23 | 88.65±1.15 | 94.15±0.24 |
| | 20 | **91.91±0.13** | 91.83±0.20 | 91.24±0.17 | 85.61±0.65 | 90.50±0.17 |
| Caltech101 | 0 | 90.29±0.00 | 88.73±0.64 | 82.39±0.48 | **90.71±0.40** | 89.48±0.28 |
| | 5 | **89.49±0.19*** | 86.52±0.37 | 73.77±0.58 | 83.21±0.88 | 84.00±0.33 |
| | 10 | **89.08±0.43*** | 85.80±0.64 | 71.58±0.85 | 81.75±0.46 | 82.92±0.98 |
| | 20 | **86.63±0.67*** | 81.46±0.16 | 66.64±1.08 | 74.59±0.53 | 79.82±0.79 |

As the results in Table 2 and Figure 10 indicate, RDML and BLMNN significantly outperform other DML methods and the performance of these two methods declines slowly than other ones with the increase of noise level. Also, RDML is consistently better than BLMNN in most of the evaluated datasets.

Other advantages of the proposed method over BLMNN include lower run time (refer to Figure 12) as well as its remarkable flexibility so that it can be easily applied to any DML algorithm based on the Hinge loss.

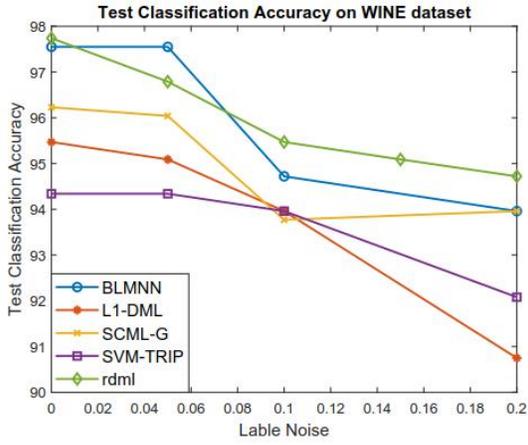

(a)

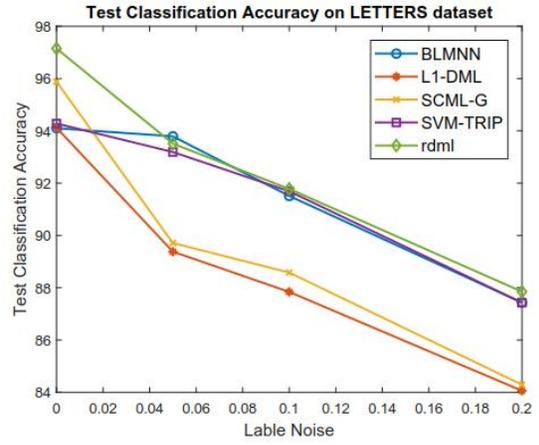

(b)

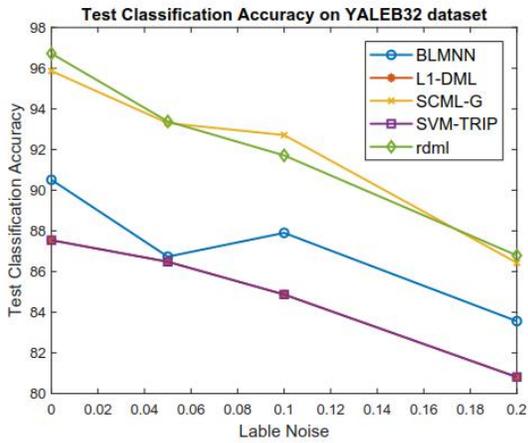

(c)

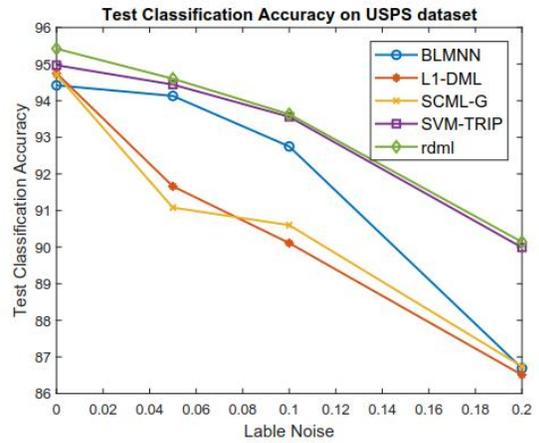

(d)

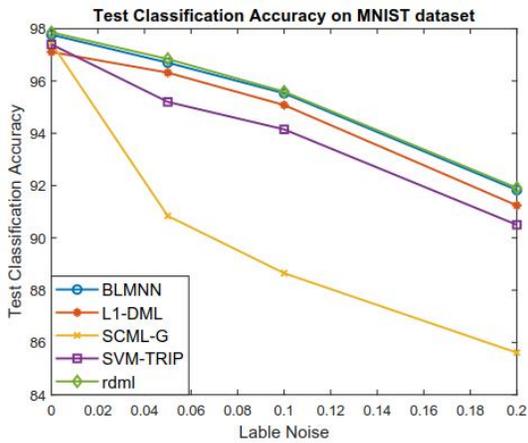

(e)

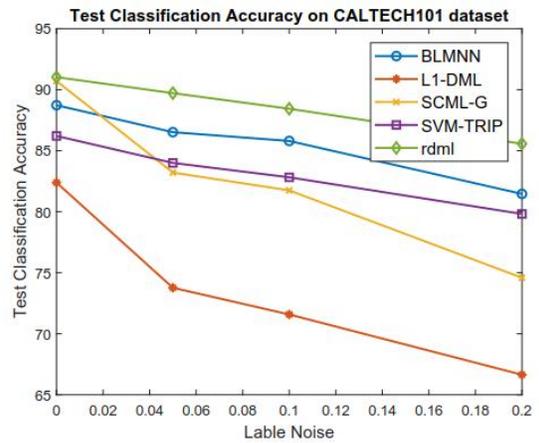

(f)

**Figure 10: Comparison of the classification accuracy of RDML with other DML methods versus label noise.**

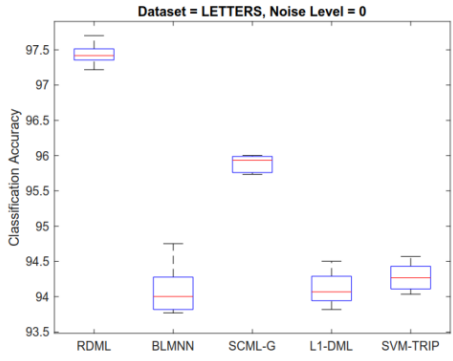

(a)

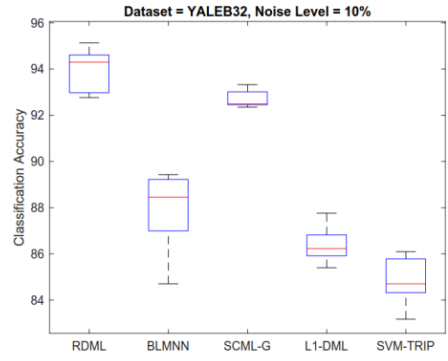

(b)

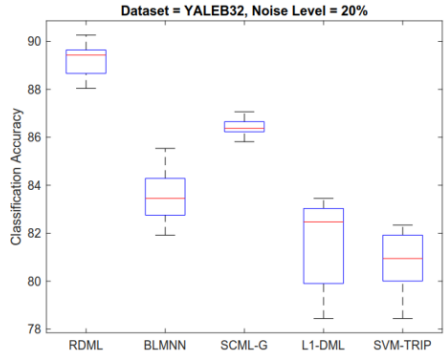

(c)

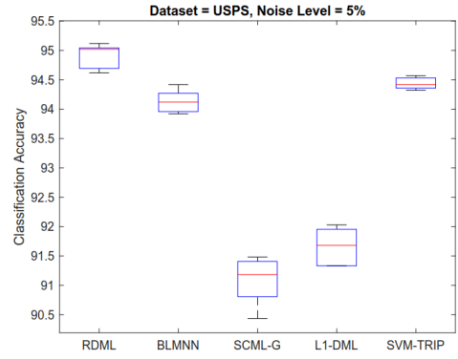

(d)

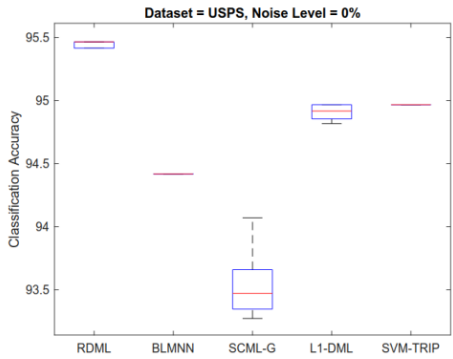

(e)

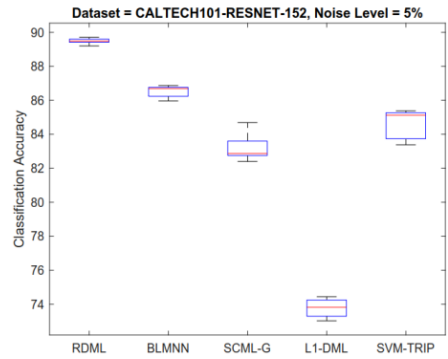

(f)

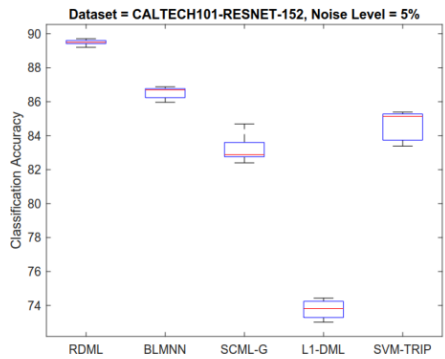

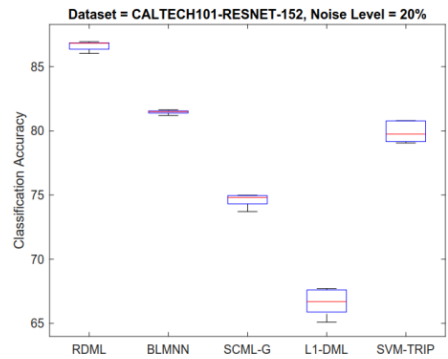

**Figure 11: Box-Plots of the results, which have statistically significant differences.**

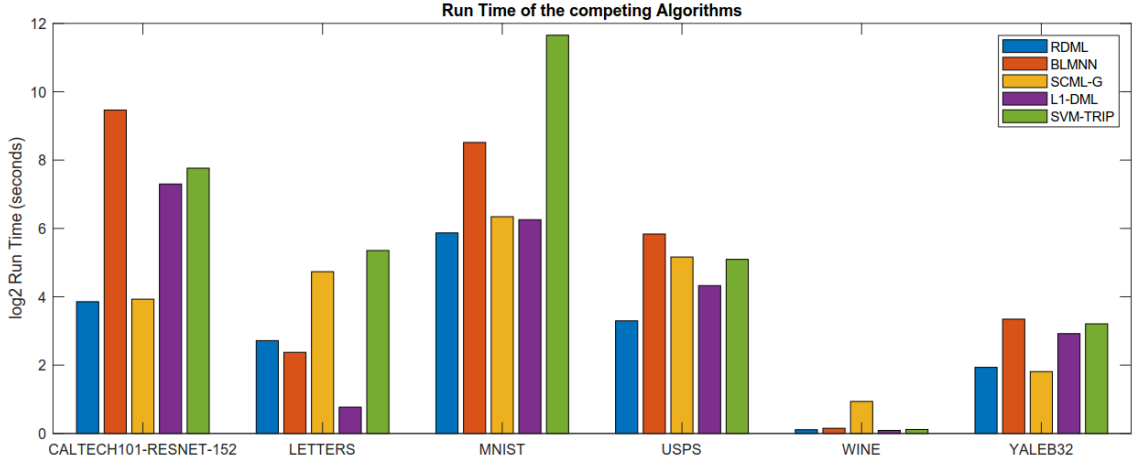

Figure 12: Comparison of the run time of RDML with other DML methods.

## 5. Conclusion and Future Work

Datasets with noisy information are common in today's applications. The main reason is the fact that many datasets are collected from the Internet using crowdsourcing or similar techniques. The wrong side information significantly diminishes the performance of DML algorithms.

To address this emerging challenge, the paper presents a robust DML method (named RDML) based on the Rescaled Hinge loss named RDML. Specifically, we initially formulate the DML problem with the robust loss function and then develop an efficient algorithm base on HQ to solve the problem. The proposed approach is rather general and one can easily apply it to any DML algorithm based on the Hinge loss.

Several experiments are conducted on both synthetic and real datasets to measure the performance of the proposed method. Experimental results indicate that RDML can effectively identify the noisy side information and reduce their influences in the metric learning process. Also, the results confirm that RDML significantly surpasses other peer DML methods in many of the evaluated datasets and its performance degrades slowly than other ones with the increase of noise level.

Some directions for future work in this area include:

I. Investigating the performance of the proposed methods in other applications like *CBIR (Content Based Information Retrieval)*.
II. Extension of the proposed method for *semi-supervised* learning.
III. Examining other robust loss functions in the proposed framework.

## Acknowledgment

We would like to acknowledge the Machine Learning Lab. in Engineering Faculty of FUM for their kind and technical support.